\documentclass[runningheads]{llncs}
\usepackage[T1]{fontenc}
\usepackage{graphicx}
\usepackage{amsmath} 
\usepackage{booktabs}
\usepackage{xcolor}
\usepackage{amsfonts}
\usepackage{mathrsfs}
\usepackage{bbm}
\usepackage{amsmath}

\usepackage{array}
\usepackage{multirow}
\usepackage{algorithm}
\usepackage{algpseudocode}

\begin{document}
\title{Confidence-Aware Automated Assessment of Student-Drawn Scientific Models
}

\titlerunning{Confidence-Aware Scoring of Student Drawings}
%


\newcommand*\samethanks[1][\value{footnote}]{\footnotemark[#1]}

\author{ Luyang Fang\inst{1,2}\orcidID{0009-0003-2465-6864} \and
Yingchuan Zhang\inst{2}\orcidID{0000-0002-4248-5128} \and
Jongchan Park\inst{1}\orcidID{0000-0002-3257-125X} \and
Zhaoji Wang\inst{1}\orcidID{0009-0007-2254-4438} \and
Ping Ma\inst{2} \orcidID{0000-0002-5728-3596} \and
Xiaoming Zhai\inst{1}\orcidID{0000-0003-4519-1931}\thanks{Corresponding author. \email{xiaoming.zhai@uga.edu} } 
}
\authorrunning{L. Fang et al.}
\institute{AI4STEM Education Center, Athens, GA, USA\\
\email{\{Luyang.Fang,Jongchan.Park,zhaojiwang,Xiaoming.Zhai\}@uga.edu}
\and
Department of Statistics, University of Georgia, Athens, GA, USA\\
\email{\{Yingchuan.Zhang,pingma\}@uga.edu}
}

%
\maketitle              

\begin{abstract}

Student-generated drawings are widely used in science education to assess learners’ conceptual understanding in modeling-based tasks aligned with the Next Generation Science Standards (NGSS). However, scoring such drawings requires expert human judgment to interpret complex visual representations, making large-scale assessment costly to implement and sustain in classroom settings.
In this work, we study automated scoring of student-generated scientific drawings using a vision-based model. We evaluate a Vision Transformer (ViT) with parameter-efficient adaptation and propose a confidence-aware scoring framework that derives response-level confidence from test-time predictive distributions. This confidence signal enables selective automation by scoring high-confidence responses automatically while deferring uncertain cases for human review.
Experiments on six NGSS-aligned middle school assessment items show that the proposed approach improves scoring reliability while supporting a practical trade-off between automated coverage and scoring risk, highlighting the value of confidence-aware methods for trustworthy educational assessment.


\end{abstract}

\section{Introduction}

In science education, assessing scientific modeling practice involves eliciting and interpreting evidence of students’ understanding through a combination of drawings and written explanations \cite{lee2023automated}. Since drawings can externalize structures, relationships, and mechanisms that may remain implicit in text-only responses, modeling assessments often rely on rubric-based evaluation of student-generated drawings alongside written explanations \cite{rahaman2024automated}. However, consistent rubric-based scoring of drawings requires careful human judgment and sustained attention to scoring quality; in everyday classroom contexts, the time, expertise, and attention required for consistent and reliable scoring are often limited \cite{lee2023automated}. At scale, reliance on human scoring becomes a major hurdle because scoring students’ drawings is labor-intensive and costly \cite{leong2018toward}. These constraints have motivated automated scoring approaches that aim to support rubric-aligned interpretation of student drawings more efficiently \cite{lee2025nerif}.


Prior research has demonstrated the feasibility of computational analysis of student-generated visual artifacts. Early work focused on extracting quantifiable visual features from drawings and linking them to rubric-aligned scoring criteria to support automated assessment \cite{leong2018toward,pei2019using}. More recently, advances in artificial intelligence have enabled modern models, including Vision Transformer (ViT) architectures \cite{han2022survey}, to directly learn scoring-relevant representations from student responses \cite{lee2025nerif,li2025utilizing,fang2026generalizable,latif2024knowledge}.


Despite these advances, existing automated scoring approaches typically produce only a single score or proficiency label, offering limited support for how such outputs should be used in instructional decision-making. In formative assessment contexts, teachers must judge not only the predicted score but also its reliability, particularly for visually complex and ambiguous student drawings. The absence of explicit confidence information makes it difficult to determine when automated scores can be trusted and when manual review is warranted, limiting the practical usability of automated assessment systems \cite{black1998assessment,gurtl2025automated}.

To address these gaps, we propose a confidence-aware \emph{vision-based} approach for automated scoring of student-generated scientific drawings. Building on the confidence-guided inference research \cite{bahat2020classification,fu2025deep,wang2019aleatoric}, the approach derives a response-level confidence score from the stability of a ViT-based model’s predictions under semantic-preserving test-time perturbations. We further incorporate \emph{test-time selection with selective trust} to emphasize reliable perturbed predictions when forming the final scoring decision. The resulting confidence score enables selective automated scoring: the system can auto-score responses with strong predictive support while deferring low-confidence cases for human review, helping teachers triage responses and calibrate reliance on automated scores in practice.

Our study evaluates the proposed framework on student-generated drawings from six NGSS-aligned middle school science modeling assessment items with rubric-based proficiency levels. Empirically, the approach improves agreement with expert scoring and provides an intuitive confidence score indicating when automated predictions are reliable, supporting informed decisions about when human review is needed in classroom assessment.

\section{Methodology}\label{sec:method}

\noindent\textbf{Problem Setup.}
We consider an automated scoring task with training data $\mathcal{D}=\left\{\left(x_i, y_i\right)\right\}_{i=1}^n$, where $x_i \in \mathcal{X}$ is a student-generated drawing represented as an image and $y_i \in\{1, \ldots, K\}$ is the corresponding rubric-based proficiency level. We train a scoring model $f_\theta: \mathcal{X} \rightarrow [0,1]^K$, where $f_\theta(x)=\left(p_1(x), \ldots, p_K(x)\right)$ is a probability distribution over the $K$ score levels.

In educational assessment settings, automated scoring systems must decide not only what score to assign, but also whether that score can be trusted. We therefore adopt a confidence-aware scoring paradigm in which the model outputs both a predicted score and a confidence value that determines whether a response is automatically scored or deferred to human review. This confidence is derived from the model’s test-time predictive distribution and is therefore directly aligned with the final scoring decision.

\vspace{6pt}
\noindent\textbf{Parameter-Efficient Task Adaptation via LoRA.}
To implement this confidence aware scoring framework in a practical classroom assessment setting, we adapt a pretrained transformer-based vision backbone using a parameter-efficient fine-tuning strategy. Specifically, we employ Low-Rank Adaptation (LoRA) \cite{hu2021lora}, which enables efficient task-specific adaptation while updating only a small subset of parameters \cite{han2024peftsurvey}.

Concretely, rather than updating all parameters in the transformer, LoRA introduces low-rank updates into the linear projection layers. For a weight matrix $W \in \mathbb{R}^{d \times p}$, the adapted layer is parameterized as
$$
W^{\prime}=W+\Delta W, \quad \Delta W=B A,
$$
where \(A \in \mathbb{R}^{r \times p}\) and \(B \in \mathbb{R}^{d \times r}\) are trainable low-rank matrices with \(r \ll \min\{d,p\}\). During training, only the LoRA parameters \(\{A,B\}\) are optimized, while the pretrained weights \(W\) remain fixed. This yields an adapted scoring function \(f_{\theta+\Delta\theta}(x)\) that captures task-specific scoring patterns with only a small number of additional trainable parameters.



\vspace{6pt}
\noindent\textbf{Confidence via Test-Time Predictive Distribution.}
Once the scoring model is adapted to the task, we estimate prediction confidence by examining the stability of scoring decisions under plausible test-time variations of the input, following the direction of \cite{bahat2020classification,fu2025deep}.

Given a response $x$, we generate $M$ semantic-preserving perturbations, such as crops or rotations,
$
\tilde{x}_j = T_j(x), j = 1, \ldots, M .
$
For each perturbed input, the model produces a predictive distribution
$p^{(j)} = f_\theta(\tilde{x}_j)$.
We then define the test-time predictive distribution as the average over these perturbations:
$
\bar{p}(x) = \frac{1}{M} \sum_{j=1}^M p^{(j)} .
$
The automated score is then given by
$
\hat{y}(x) = \arg\max_k \bar{p}_k(x).
$

We define the \emph{response-level confidence score} as the probability mass
assigned to the predicted score:
\[
\kappa(x) = \max_k \bar{p}_k(x)
          = \bar{p}_{\hat{y}(x)}(x)
          = \frac{1}{M} \sum_{j=1}^M p^{(j)}_{\hat{y}(x)} .
\]
Here, $\kappa(x) \in [0,1]$ measures how strongly the final predictive distribution supports the chosen score. Intuitively, a high confidence value indicates that plausible test-time variations of the response largely agree on the same score, whereas a low value
suggests disagreement among these variations.


\vspace{6pt}
\noindent\textbf{Selective Automated Scoring.}
Using the response-level confidence score defined above, we implement a selective automated scoring strategy that defers low-confidence cases to human graders. Given a confidence threshold $\tau$, define
\(
g_\tau(x) = \mathbf{1}\{\kappa(x) \ge \tau\}.
\)
The selective scoring rule is:
\[
h_\tau(x) =
\begin{cases}
\hat{y}(x), & g_\tau(x) = 1, \\
\text{defer to human review}, & g_\tau(x) = 0.
\end{cases}
\]
This formulation allows the system to automatically score responses with sufficient confidence while deferring uncertain cases for manual evaluation. By varying $\tau$, we control the trade-off between automated coverage and scoring risk.

\vspace{6pt}
\noindent\textbf{Test-Time Selection with Selective Trust.}
While the test-time predictive distribution already provides a response-level confidence score, we further improve robustness by selectively trusting individual test-time predictions. Importantly, this step operates at the \emph{view level} and uses an internal \emph{selection score}, which is distinct from the response-level confidence $\kappa(x)$ defined above.

For each augmented prediction $p^{(j)}$, let $\hat{y}^{(j)} = \arg\max_k p^{(j)}_k$ denote the predicted class under that perturbation. We then compute a selection score:
\begin{equation}
c^{(j)} = -\frac{1}{K-1} \sum_{k \neq \hat{y}^{(j)}} \log \big(p^{(j)}_k \big),
\end{equation}
which measures how strongly that perturbed prediction suppresses probability mass on competing classes. Larger values of $c^{(j)}$ indicate more decisive predictions for that test-time view and are used solely for ranking and selecting reliable perturbations.

We adopt a Top-$\eta$ filtering strategy at test time. Given $M$ perturbed views of an input $x$, we retain only the top $\lceil \eta M \rceil$ perturbations with the largest selection scores $c^{(j)}$, where $\eta \in (0,1]$. Let $\mathcal{J}(x) \subseteq \{1,\ldots,M\}$ denote the indices of these selected views. The filtered predictive distribution is then defined as
\[
\bar{p}_\eta(x) = \frac{1}{|\mathcal{J}(x)|} \sum_{j \in \mathcal{J}(x)} p^{(j)} .
\]
The refined prediction and response-level confidence are given by
\[
\hat{y}_\eta(x) = \arg\max_k \bar{p}_{\eta,k}(x),
\quad
\kappa_\eta(x) = \max_k \bar{p}_{\eta,k}(x).
\]



\section{Dataset Details}\label{sec:data}

We use a dataset of student-generated scientific drawings collected from middle school science modeling assessments in the northeastern United States \cite{harris2024creating}. The assessment items are aligned with the Next Generation Science Standards (NGSS) and require students to construct visual models to explain observed scientific phenomena.
Our experiments focus on the visual components of student responses from six assessment items, each targeting a distinct scientific concept. All responses are provided as images and independently scored by domain experts using rubric-based criteria. Following the original annotation protocol, each drawing is assigned to one of three ordered proficiency levels:
\textit{Beginning}, \textit{Developing}, or \textit{Proficient}.

\begin{table}[t]
\centering
\caption{Summary of the student drawing dataset used in this study.}
\label{tab:dataset_stats}
\scalebox{0.85}{%
\begin{tabular}{
>{\centering\arraybackslash}m{1.3cm}
>{\centering\arraybackslash}m{3.2cm}
>{\centering\arraybackslash}m{1.1cm}
>{\centering\arraybackslash}m{1.9cm}
>{\centering\arraybackslash}m{1.9cm}
>{\centering\arraybackslash}m{1.9cm}
}
\toprule
\textbf{Item} & \textbf{Item Description} & \textbf{Total} & \textbf{Beginning} & \textbf{Developing} & \textbf{Proficient} \\
\midrule
Item 1 & Red dye diffusion        & 477 & 195 & 205 & 77  \\
Item 2 & Jane’s inflated ball     & 538 & 177 & 288 & 73  \\
Item 3 & Melting butter           & 520 & 155 & 266 & 99  \\
Item 4 & Hot shower effect        & 772 & 494 & 107 & 171 \\
Item 5 & Heated cup of water      & 453 & 61  & 262 & 130 \\
Item 6 & Jennifer’s teapot        & 816 & 390 & 271 & 155 \\
\bottomrule
\end{tabular}}
\end{table}

\begin{figure}[t]
\centering
\includegraphics[width=0.75\linewidth]{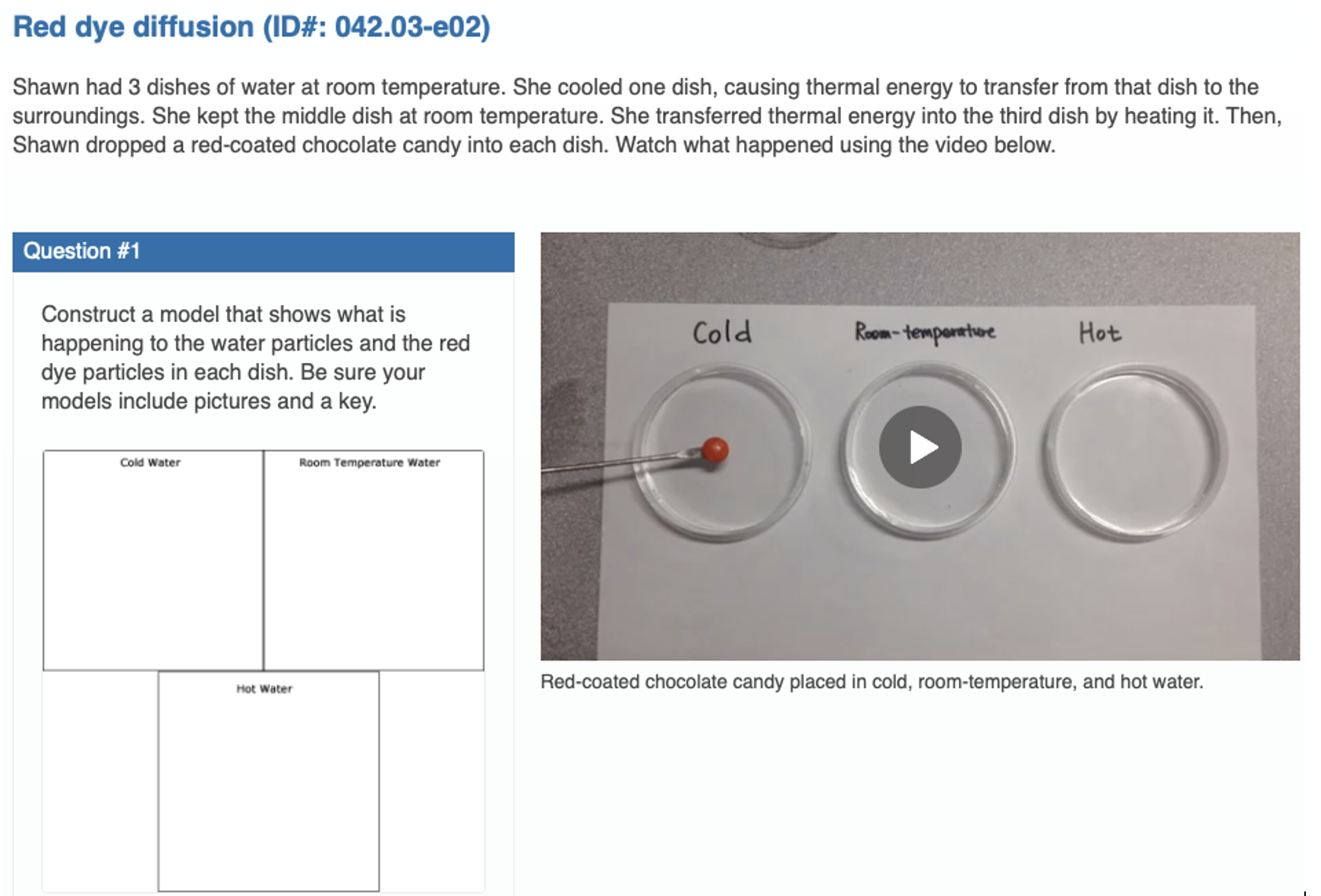}
\caption{Example science modeling assessment item. Students observe red dye diffusion in cold, room-temperature, and hot water and are asked to construct a visual model representing the behavior of water and dye particles.}\label{fig:example_item}
\end{figure}

Figure~\ref{fig:example_item} shows an example item illustrating the open-ended and visually complex nature of the student drawings. Example student drawings are available in the GitHub repository \footnote{\url{https://github.com/LuyangFang/CA-Drawing}}.
Dataset statistics for each item, including the number of responses and label distributions, are summarized in Table~\ref{tab:dataset_stats}.



\vspace{-5pt}


\section{Experimentation and Results}\label{sec:exp}

We evaluate the proposed confidence-aware automated scoring framework on the student drawing assessment dataset described in Section \ref{sec:data}. For each assessment item, the task is formulated as a supervised multi-class classification problem, where the scoring model predicts one of $K=3$ rubric-aligned proficiency levels: Beginning, Developing, or Proficient.

\noindent\textbf{Baselines.}
We compare four approaches: (1) \textbf{ViT (Frozen)}, a pretrained vision transformer without task-specific fine-tuning; (2) \textbf{ViT+LoRA}, where the backbone is adapted using low-rank adaptation; (3) \textbf{CA-Uniform}, a confidence-aware scoring method that uniformly trusts all test-time predictions; and (4) \textbf{CA-Selective}, our proposed confidence-aware selective scoring method
based on test-time selection with selective trust.

\noindent\textbf{Evaluation Metrics.}
We report performance using (1) standard classification metrics, including accuracy, Cohen’s $\kappa$, precision, recall, and F1 score, computed separately for each item and averaged across items; and (2) evaluation of the confidence score introduced by the proposed CA-Selective method.


\noindent\textbf{Training Details.}
Models are trained using a stratified train/validation/test split to preserve proficiency distributions. We adopt a pretrained ViT model (\texttt{vit\_base\_patch16\_224}) backbone. Standard image preprocessing and data augmentation are applied. Models are optimized using cross-entropy loss for three-class classification, and the best checkpoint for each item is selected based on validation Cohen’s $\kappa$. We use $M=20$ and $\eta=0.75$.
All models are trained independently for each assessment item.

\vspace{-10pt}
\begin{table}[h]
\centering
\caption{Comparison of four methods across six items. Best performance for each item is shown in bold.}
\label{tab:method_comparison}
\scalebox{0.9}{%
\begin{tabular}{
>{\centering\arraybackslash}m{1.5cm}
>{\centering\arraybackslash}m{2.5cm}
>{\centering\arraybackslash}m{1.2cm}
>{\centering\arraybackslash}m{1.2cm}
>{\centering\arraybackslash}m{1.2cm}
>{\centering\arraybackslash}m{1.2cm}
>{\centering\arraybackslash}m{1.2cm}
>{\centering\arraybackslash}m{1.2cm}
>{\centering\arraybackslash}m{1.2cm}
}
\toprule
\textbf{Metric} & \textbf{Method} & \textbf{Item 1} & \textbf{Item 2} & \textbf{Item 3} & \textbf{Item 4} & \textbf{Item 5} & \textbf{Item 6} & \textbf{Avg} \\
\midrule[\heavyrulewidth]

\multirow{4}{*}{Accuracy}
 & ViT (Frozen) & 0.438 & 0.482 & 0.212 & 0.244 & 0.174 & 0.183 & 0.289 \\
 & ViT + LoRA & \textbf{0.792} & 0.833 & 0.769 & \textbf{0.821} & 0.761 & 0.622 & 0.766 \\
 & CA-Uniform & \textbf{0.792} & \textbf{0.852} & 0.808 & \textbf{0.821} & \textbf{0.783} & \textbf{0.659} & 0.786 \\
 & CA-Selective & \textbf{0.792} & \textbf{0.852} & \textbf{0.827} & \textbf{0.821} & \textbf{0.783} & \textbf{0.659} & \textbf{0.789} \\
\midrule

\multirow{4}{*}{Kappa}
 & ViT (Frozen) & -0.048 & 0.134 & -0.110 & 0.040 & -0.112 & 0.000 & -0.016 \\
 & ViT + LoRA & \textbf{0.796} & 0.759 & 0.708 & 0.788 & 0.589 & 0.608 & 0.708 \\
 & CA-Uniform & \textbf{0.796} & \textbf{0.789} & 0.747 & \textbf{0.813} & 0.610 & \textbf{0.643} & 0.733 \\
 & CA-Selective & \textbf{0.796} & \textbf{0.789} & \textbf{0.826} & \textbf{0.813} & \textbf{0.694} & \textbf{0.643} & \textbf{0.760} \\
\midrule

\multirow{4}{*}{Precision}
 & ViT (Frozen) & 0.314 & 0.304 & 0.246 & 0.411 & 0.258 & 0.061 & 0.266 \\
 & ViT + LoRA & \textbf{0.774} & 0.905 & 0.728 & \textbf{0.721} & 0.813 & 0.585 & 0.754 \\
 & CA-Uniform & \textbf{0.774} & \textbf{0.914} & 0.774 & 0.708 & 0.829 & \textbf{0.638} & 0.773 \\
 & CA-Selective & \textbf{0.774} & \textbf{0.914} & \textbf{0.805} & 0.708 & \textbf{0.836} & \textbf{0.638} & \textbf{0.779} \\
\midrule

\multirow{4}{*}{Recall}
 & ViT (Frozen) & 0.350 & 0.327 & 0.259 & 0.347 & 0.255 & 0.333 & 0.312 \\
 & ViT + LoRA & \textbf{0.758} & 0.695 & 0.736 & \textbf{0.702} & 0.695 & 0.596 & 0.697 \\
 & CA-Uniform & \textbf{0.758} & \textbf{0.713} & 0.782 & 0.679 & \textbf{0.707} & \textbf{0.639} & 0.713 \\
 & CA-Selective & \textbf{0.758} & \textbf{0.713} & \textbf{0.815} & 0.679 & \textbf{0.707} & \textbf{0.639} & \textbf{0.719} \\
\midrule

\multirow{4}{*}{F1 Score}
 & ViT (Frozen) & 0.257 & 0.299 & 0.196 & 0.152 & 0.173 & 0.103 & 0.196 \\
 & ViT + LoRA & \textbf{0.765} & 0.729 & 0.731 & \textbf{0.708} & 0.724 & 0.571 & 0.705 \\
 & CA-Uniform & \textbf{0.765} & \textbf{0.745} & 0.778 & 0.687 & 0.739 & \textbf{0.612} & 0.721 \\
 & CA-Selective & \textbf{0.765} & \textbf{0.745} & \textbf{0.809} & 0.687 & \textbf{0.743} & \textbf{0.612} & \textbf{0.727} \\
\bottomrule
\end{tabular}}
\end{table}

\vspace{-10pt}
\subsection{Automated Scoring Performance}

Table \ref{tab:method_comparison} reports automated scoring performance across six NGSS-aligned modeling items. Overall, the confidence-aware methods improve average performance relative to standard fine-tuning, with the clearest gains observed in Cohen’s kappa and, in most cases, F1 score. Higher kappa values indicate improved agreement with human rubric-based scoring beyond chance, suggesting that confidence-aware modeling can support more reliable scoring decisions across diverse student drawings.


Comparing the two confidence-aware variants, CA-Selective achieves the best overall performance on average across items, suggesting that selective aggregation can improve scoring quality by reducing the influence of less informative test-time predictions. From an assessment perspective, this leads to rubric-aligned predictions that are less sensitive to superficial variability in students’ representational styles. Such robustness is especially important in open-ended drawing tasks, where similar ideas may be expressed visually in different ways.



Table \ref{tab:efficiency} summarizes model complexity across scoring methods. All approaches share the same ViT backbone (86.4M parameters), so performance differences are not attributable to model capacity. LoRA introduces only 0.6M additional trainable parameters while keeping the backbone fixed, and the confidence-aware variants add no further trainable parameters. Confidence-aware scoring incurs a higher inference cost because it aggregates multiple test-time augmented views to obtain response-level confidence, but all methods remain substantially lighter than LLM-based scoring approaches.

\vspace{-10pt}
\begin{table}[h]
\centering
\caption{Model complexity and computational cost of different scoring methods.}
\label{tab:efficiency}
\scalebox{0.9}{%
\begin{tabular}{
>{\centering\arraybackslash}m{2.8cm}
>{\centering\arraybackslash}m{2.8cm}
>{\centering\arraybackslash}m{2.8cm}
>{\centering\arraybackslash}m{2.8cm}
}
\toprule
Method & Backbone Params & Trainable Params & Inference Latency (ms) \\
\midrule
ViT (Frozen) & 86.4M & 0 & 1.0341 \\
ViT + LoRA & 86.4M & 0.6M & 1.0355 \\
CA-Uniform & 86.4M & 0.6M & 20.532 \\
CA-Selective & 86.4M & 0.6M & 20.572 \\
\bottomrule
\end{tabular}}
\end{table}

To provide comparison with modern open-source multimodal foundation models, we additionally evaluated a vision–language model (Qwen3-VL-8B-Instruct) in a zero-shot setting using the same test split and rubric-aligned prompts. In this configuration, the model achieved lower agreement with expert scoring and substantially higher inference latency than the task-adapted ViT-based approaches. Details are provided in the GitHub repository.
This does not indicate a limitation of the multimodal model itself, as effective use of VLMs for rubric-aligned assessment typically benefits from careful prompt design, few-shot exemplars, or task-specific adaptation. Instead, the comparison provides a practical reference showing that lightweight task-adapted vision backbones remain competitive and efficient in this setting.

\subsection{Confidence Scores}

To validate the intuitive rationality of our confidence metric $\kappa(x)$, we analyze the correlation between mean confidence and predictive accuracy across all proficiency labels. 
We find that there is a significant positive correlation ($r = 0.649, p < 0.01$) between the two. The linear fit demonstrates that higher confidence scores consistently align with higher scoring accuracy, particularly for the `Beginning' level. This alignment ensures that $\kappa(x)$ serves as a reliable proxy for scoring quality, providing a solid foundation for the subsequent selective automated scoring strategy.
Additional visualization results are available in the GitHub repository. We are also conducting expert review to verify the qualitative validity of these results.

In assessment contexts, the key question is not only whether a model can generate a score, but whether that score should be acted upon without further review. In this regard, the confidence score may provide a practical basis for deciding when an automated score may be used and when teacher review is still needed. This is especially relevant for student drawings, where visually ambiguous or unconventional drawings may be difficult for automated systems to interpret reliably and therefore may require teacher review.


\section{Conclusion}

In this work, we studied confidence-aware automated scoring of student-generated scientific drawings using a vision-based approach. By deriving response-level confidence from test-time predictive distributions and selectively aggregating reliable predictions, the proposed framework enables automated scoring that balances efficiency and reliability. Experiments on six NGSS-aligned science modeling tasks demonstrate that confidence-aware methods improve agreement with human scoring while providing an intuitive signal for deciding when automated scores can be trusted. Beyond performance gains, this work highlights the importance of integrating confidence estimation into automated assessment systems to support responsible deployment in classroom settings. At the same time, findings in this study should be interpreted in light of the dataset context. The data were collected from middle school science classrooms in one region of the United States, and students' representational practices may reflect local curricular and classroom contexts. In addition, because expert-provided scores were used as the reference, any systematic tendencies in human scoring may also be reflected in model performance. With these caveats, future work will examine the proposed approach across more diverse student populations and assessment contexts, as well as explore extensions to multimodal student responses and more fine-grained feedback aligned with instructional use.

\begin{credits}
\subsubsection{\ackname} 
This work was partially supported by the U.S. National Science Foundation (NSF) [2101104,  DMS-2124493, DMS-2311297, DMS-2319279, DMS-2318809]. 
Any opinions, findings, conclusions, or recommendations expressed in this material are those of the authors and do not necessarily reflect the views of NSF.

\subsubsection{\discintname}
The authors have no competing interests to declare.
\end{credits}

\bibliographystyle{splncs04}
\bibliography{ref}
%






\end{document}